\documentclass[letterpaper, 10 pt, journal, twoside]{IEEEtran}
\usepackage{cite}
\usepackage{array}
\usepackage{float}
\usepackage{xcolor}
\usepackage{lipsum}
\usepackage{caption}
\usepackage{wrapfig}
\usepackage{graphicx}
\usepackage{textcomp}
\usepackage{fixltx2e}
\usepackage{hyperref}
\usepackage{subcaption}
\usepackage{algorithmic}
\usepackage{multirow}
\usepackage{tabularx}
\usepackage[euler]{textgreek}
\usepackage[affil-it]{authblk}
\usepackage{amsmath,amssymb,amsfonts}
\usepackage{enumerate}
\DeclareMathOperator*{\argmin}{argmin}

\author{Shilin Shan$^{1,*}$ and Quang-Cuong Pham$^{2}$ \vspace{-20pt}
\thanks{Manuscript received: April, 18, 2024; Revised July, 11, 2024; Accepted August, 12, 2024.}
\thanks{This research was supported by the National Research Foundation, Prime Minister's Office, Singapore under its Medium Sized Centre funding scheme, Singapore Centre for 3D Printing, CES\_SDC Pte Ltd, and Chip Eng Seng Corporation Ltd.}
\thanks{$^{1}$ Shilin Shan is with School of Mechanical and Aerospace Engineering, Nanyang Technological University, Singapore (address: 50 Nanyang Ave, Singapore 639798; phone: +65 6790 5568; e-mail: shilin.shan153@gmail.com)} 
\thanks{$^{2}$ Quang-Cuong Pham is with Eureka Robotics and Singapore Centre for 3D Printing (SC3DP), School of Mechanical and Aerospace Engineering, Nanyang Technological University, Singapore. (e-mail: cuong@ntu.edu.sg)} 
\thanks{$^{*}$ Corresponding author: Shilin Shan} 
}

\title{Fast Payload Calibration for Sensorless Contact Estimation Using Model Pre-training} 

\begin{document}

\maketitle

\begin{abstract}

    Force and torque sensing is crucial in robotic manipulation across both collaborative and industrial settings. Traditional methods for dynamics identification enable the detection and control of external forces and torques without the need for costly sensors. However, these approaches show limitations in scenarios where robot dynamics, particularly the end-effector payload, are subject to changes. Moreover, existing calibration techniques face trade-offs between efficiency and accuracy due to concerns over joint space coverage. In this paper, we introduce a calibration scheme that leverages pre-trained Neural Network models to learn calibrated dynamics across a wide range of joint space in advance. This offline learning strategy significantly reduces the need for online data collection, whether for selection of the optimal model or identification of payload features, necessitating merely a 4-second trajectory for online calibration. This method is particularly effective in tasks that require frequent dynamics recalibration for precise contact estimation. We further demonstrate the efficacy of this approach through applications in sensorless joint and task compliance, accounting for payload variability.

\end{abstract}

\begin{IEEEkeywords}
    Machine Learning for Robot Control, Industrial Robots, Physical Human-Robot Interaction
\end{IEEEkeywords}

\vspace{-4pt}
\section{Introduction}

\IEEEPARstart{R}{obot} manipulators are widely deployed around the globe, making force/torque (F/T) sensing crucial for various tasks. Typically, this capability is facilitated by additional 6-axis F/T sensors or joint torque sensors. However, these sensors are expensive and may be undesirable in certain situations due to their weight and volume. Consequently, researchers have proposed sensorless contact estimation techniques that rely on dynamics identification, either model-based \cite{de2006collision, wahrburg2017motor} or model-free \cite{nguyen2008local, yilmaz2020neural}. These techniques estimate external contacts using proprioceptive signals, such as motor current and joint states.

However, many of these approaches overlook scenarios where robot dynamics are subject to change. To address this, various methods have been proposed to mitigate errors resulting from payload variations \cite{gaz2017payload, dong2018efficient, selingue2023hybrid}. These solutions typically involve a trade-off between achieving high accuracy across a wide range of joint space, determined by the volume of calibration data, and the time required for calibration trajectories. However, both precision and efficiency are essential for many collaborative tasks, especially when the robot must quickly grasp and assemble components of varying weights while accurately detecting external contacts with human operators and the environment.

\begin{figure}[t]
    \vspace{4pt}
    \centering
    \includegraphics[width=0.45\textwidth]{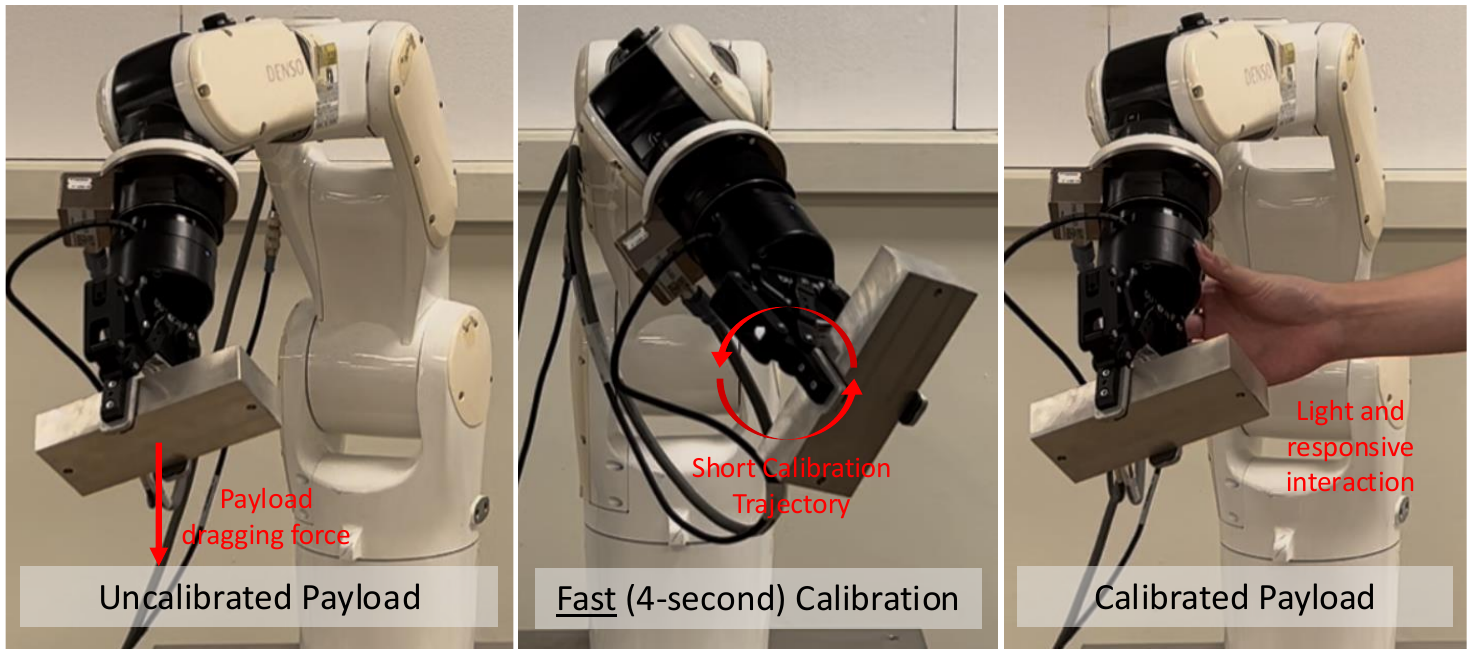}
    \caption{Snapshot of the robot compliance task. A video demonstration is available as the supplementary or at: \href{https://youtu.be/r5SYtfSI6uc}{https://youtu.be/r5SYtfSI6uc}}
    \label{Fig:Snap}
    \vspace{-8pt}
\end{figure}

To address these challenges, we introduce a fast online calibration scheme enabled by pre-trained Neural Network (NN) models. Specifically, we present both Payload-specific and Payload-adaptive model architectures. Training sets for various payload types are collected through offline trajectories that encompass the entire joint space. These NN models are then trained through offline learning, enabling the acquisition of minimal online data via a short 4-second calibration trajectory specifically aimed at identifying key payload features. This advancement significantly expands the range of applications for sensorless contact estimation techniques in environments with frequent dynamics variations. We further integrate the proposed calibration scheme into two tasks: joint space compliance and task space compliance, to demonstrate its effectiveness in real-world applications. A snapshot of the task space compliance experiment is depicted in Fig. \ref{Fig:Snap}.

This paper is structured as follows: Section II presents a literature review and discussion of calibration methods. Section III outlines the baseline model-based algorithm, online learning algorithm and the proposed offline learning schemes. Section IV details the training data collection process and the rationale behind the online calibration trajectory design. Section V compares the results obtained from all discussed methods. Finally, Section VI demonstrates the integration of the proposed method into interactive tasks.

\section{Related Work}

The study of dynamics identification has laid a solid foundation for sensorless external contact detection and estimation. In the literature, major solutions to this challenge are categorized into three groups: model-based methods \cite{de2006collision, wahrburg2017motor}, model-free methods \cite{yilmaz2020neural, sharkawy2020human}, and hybrid methods \cite{kim2021transferable}. These approaches utilize parametric dynamics modeling, black-box learning algorithms, or a combination of both, estimating external contact through measurements of motor torque, current, and joint states. Relying on current introduces additional challenges. Studies \cite{gautier2014global, xu2022robot} report that current signals tend to be noisier than torque measurements, and motor constants provided by manufacturers can be imprecise for certain robotic models. Additional efforts may be required to obtain reliable torque estimation from current signals. Despite careful signal processing, these studies commonly face another limitation: they are often conducted across a wide range of joint spaces, aiming to identify the complete robot model under a consistent robot dynamics setting.  If frequent changes in the end-effector payload necessitate calibration for external contact estimation, the standard identification process may become impractically time-consuming.

To tackle this problem, model-based methods have been developed to facilitate fast calibration by compensating for variations in the end-effector payload. In \cite{nadeau2022fast}, it is demonstrated that a 1.5-second calibration trajectory is sufficient given F/T measurements and knowledge of the payload shape, suggesting that fast payload calibration is achievable with appropriate data modality. In \cite{dong2018efficient}, the Least Squares (LS) algorithm is employed to determine the optimal set of parameters based on data collected online. Whereas, the reported calibration times remain significant. \cite{kurdas2022online} proposes gathering payload information and solving for payload parameters through similar LS algorithms by executing two similar calibration trajectories. This design allows for compensation of unmodeled disturbances with flexibility. However, a common challenge of payload estimation using proprioceptive sensors is the lack of global accuracy with calibration data collected locally. This could be caused by physically inconsistent dynamics parameters being identified using local data. \cite{gaz2017payload} summarizes that accurate estimation requires global data, but online data collection across the entire joint space remains impractical due to extensive time requirements. 

Conversely, learning-based methods have demonstrated a potent ability to capture the robot dynamics, enabling minimal calibration data use with an offline-trained model. In \cite{shareef2016generalizing}, a detailed comparison among various learning-based techniques, including Gaussian Process Regression (GPR), Echo State Networks (ESN), and Extreme Learning Machine (ELM), is provided. This comparison suggests that ELM --- a single-layer Neural Network (NN) --- outperforms other methods in tasks related to payload calibration. Further research in \cite{selingue2023hybrid} introduces a calibration strategy that trains models with various payloads. This approach allows for the estimation of joint-level compensations by interpolating between the models' predictions. However, for effective interpolation, the payloads' mass must be known beforehand; without this information, online collection of fine-tuning data across the joint space would lead to the same problem of excessive time consumption. 

In this paper, we adopt a similar model pre-training approach but introduce two different strategies to address the challenges: (i) We implement fine-grained discretization of payloads during the training phase and train payload-specific models to ensure there is always a model capable of compensating for the unknown payload with acceptable errors. (ii) We identify and incorporate payload features during the training phase, allowing for their extraction from short online trajectories. This process informs the pre-trained model about payload variations.

As a preliminary step to payload compensation, a base model is needed to estimate joint currents in a payload-free scenario. A previous study introduced a Neural-Network-based dynamics identification approach \cite{shan2024sensorless}, utilizing a specialized Motion Discriminator (MD) input scheme. This method incorporates both instantaneous joint states and temporal information as inputs, effectively reducing significant errors caused by friction hysteresis. We adopted this design to train the base model and provide estimations for the payload-free dynamics. 

\section{Calibration Methods}

In this section, we introduce four candidate calibration methods: the model-based calibration method, the Online Learning Model, the Payload-specific Pre-trained Model, and the Payload-adaptive Pre-trained Model. The architecture for the base model and all calibration models is Multilayer Perceptron (MLP). According to a previous study \cite{10354486}, MLP has been shown to outperform other architectures, including Convolutional Neural Networks (CNN) and Recurrent Neural Networks (RNN), in terms of accuracy and inference speed.

\subsection{Model-based Method}
\vspace{-3pt}
In this paper, we consider and implement the model-based calibration method discussed in \cite{gaz2017payload}, based on the robot's fundamental dynamics parameters obtained using \cite{gaz2019dynamic,shan2024sensorless}. The end-effector payload can be regarded as an extension of the robot's last link, allowing for identification of the payload parameters through modified dynamics. Specifically, the joint-level torque variation at the $n^{th}$ data frame due to payloads can be written as follows:
\begin{equation}
    Y(q_n,\dot{q}_n,\ddot{q}_n)(\pi_L - \pi) = \Delta\tau_n = \tau_{L,n} - \tau_n
\end{equation}
where $Y$, $\pi$, $\pi_L$, $\tau$, and $\tau_L$ are the regressor matrix of the robot, original dynamics parameters, loaded dynamics parameters, load-free motor torque, and loaded motor torque, respectively. By collecting calibration data online, the variation of dynamics parameters can be evaluated using the stacked formulation:
\begin{equation}
    \hat{\epsilon} = \overline{Y}^\#_L\overline{\Delta\tau}
\end{equation}
where $\hat{\epsilon}$ is the parameter variations due to end-effector payloads. Furthermore, the payload parameters:
\vspace{-0pt}
\begin{equation}
    p_L = (m_L, c_{Lx}m_L, c_{Ly}m_L, c_{Lz}m_L, J_L)^T \in \mathbb{R}^{10}
\end{equation}
can be solved explicitly using:
\vspace{-0pt}
\begin{equation}
    \hat{p}_L = J^\#_{\epsilon}\hat{\epsilon} \qquad J_{\epsilon} = \frac{\partial \epsilon(p_L)}{\partial p_L} \qquad \epsilon(p_L) = \pi_L - \pi
\end{equation}
By modifying $\pi$ using the identified payload parameters, the estimated motor torque would include the component for payload compensation. 

\subsection{Online Learning Model (OLM)}

As discussed in Section II, online learning methods are widely employed for dynamics and kinematics calibration purposes, as referenced in previous works \cite{selingue2023hybrid}. However, these methods suffer from issues such as prolonged calibration times and diminished accuracy over large workspaces. We implemented an Online Learning method as a baseline to substantiate these statements through experimental observations.

\begin{figure}[t]
    \centering
    \includegraphics[width=0.35\textwidth]{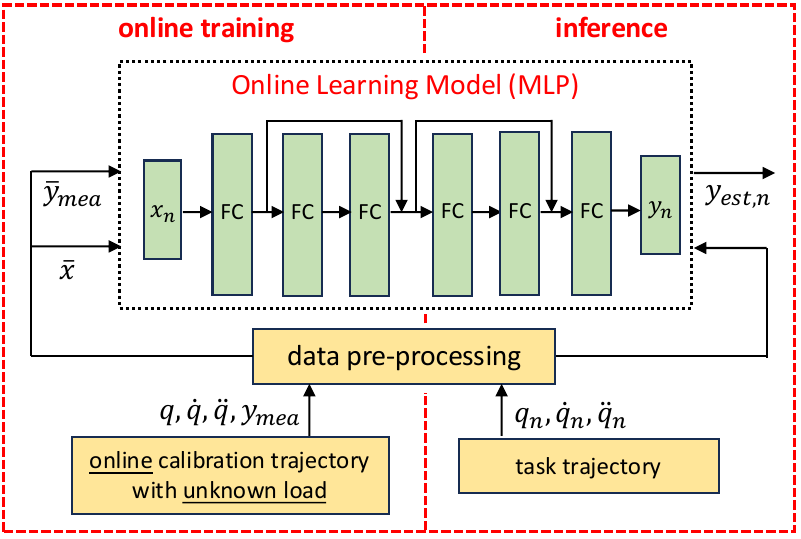}
    \caption{The block diagram and data flow for OLM. The implication of the variables can be found in Table \ref{Table:Var_Imp}.}
    \label{Fig:OnLD}
    \vspace{-10pt}
\end{figure}

We initialize the OLM using a payload-free base model that was pre-trained, as discussed in previous research \cite{shan2024sensorless}. Calibration data are then collected through short calibration trajectories, followed by the fine-tuning of the base model using Batch Gradient Descent (BGD). The base model features a 6-layer MLP architecture with residual connections. Its structure, along with the data flow of online training and inference, is depicted in Fig. \ref{Fig:OnLD}. Consistent with the training framework of the base model, we adopt the same MD input scheme to effectively reduce hysteresis error. The specifics of the input scheme are not detailed in this paper and are represented by the `data pre-processing' unit in Fig. \ref{Fig:OnLD}.

\subsection{Payload-specific Pre-trained Models (PsPM)}

As highlighted in Section II, the pre-trained model can learn the complete robot dynamics from offline data collected across the entire workspace. This approach significantly reduces the need for extensive data collection during the online calibration of an unknown payload. In this section, we introduce a straightforward yet effective method that trains distinct models for different end-effector payloads.

\begin{figure}[t]
    \centering
    \includegraphics[width=0.45\textwidth]{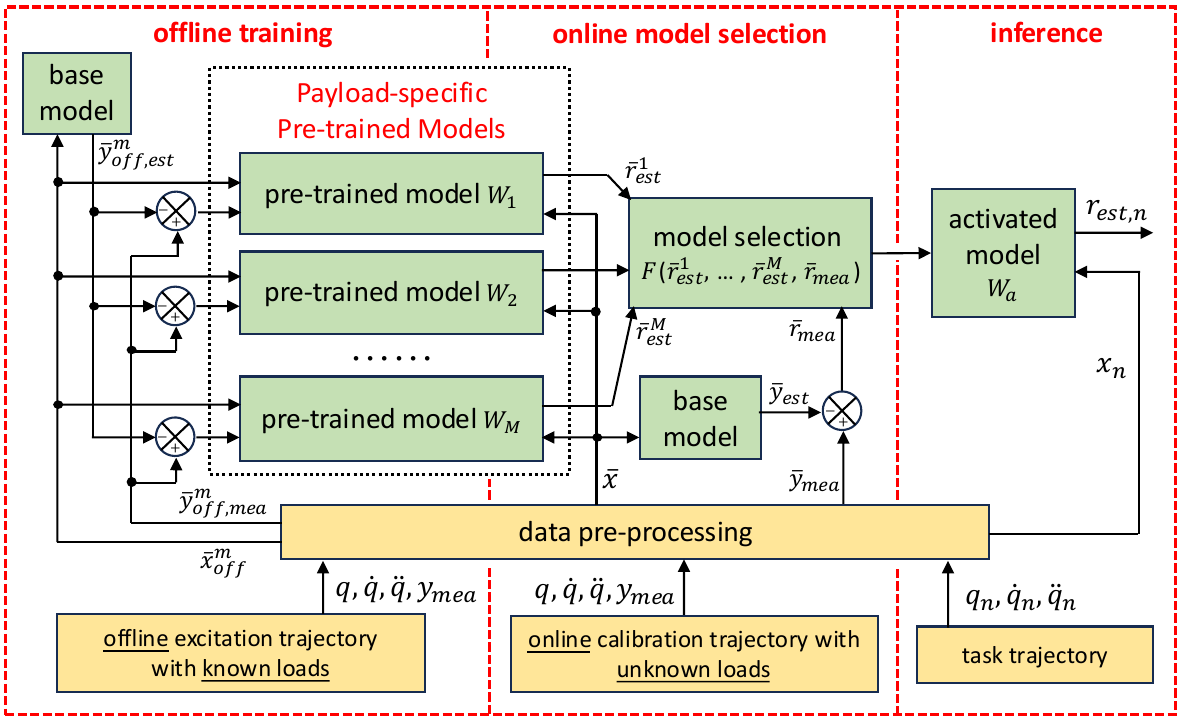}
    \caption{The block diagram and data flow for PsPM. The implication of the variables can be found in Table \ref{Table:Var_Imp}.}
    \label{Fig:OffLD}
    \vspace{-10pt}
\end{figure}

The block diagrams of data flow, model training, and online selection are illustrated in Fig. \ref{Fig:OffLD}. During the offline training phase, all models are initialized with random weights and trained to predict the current residuals associated with payloads, which are referred to as 'payload residuals' throughout this paper. Each model employs a 3-layer fully-connected (FC) MLP architecture. For training a single payload-specific model, a distinct combination of mass and center of mass (CoM) is used for data collection. This necessitates the discretization of both variables, creating finite sets of combinations. Since data collection and model training are conducted offline, it is feasible to gather extensive training data, ensuring comprehensive coverage of the entire joint space. Section IV.A will detail the required dataset size for each payload and the discretization methodology.

For the online model selection process, data are gathered from online calibration trajectories. We then obtain estimations from all models via multi-process computation and select the most accurate model based on the current estimations and measurements. Specifically, the most accurate model is selected using the following equations:

\vspace{-7pt}
\small
\begin{equation}
    W_a = F(\bar{r}^1_{est}....,\bar{r}^M_{est},\bar{r}_{mea}) = \argmin_{W_m} \bigl(\text{MSE}(\bar{r}_{mea},\bar{r}^m_{est})\bigr)
\end{equation}
\vspace{-7pt}
\begin{equation}
    \text{MSE}(\bar{r}_{mea},\bar{r}^m_{est}) = \frac{\sum_{i=1}^{I} (\bar{r}_{mea,i} - \bar{r}^m_{est,i})(\bar{r}_{mea,i} - \bar{r}^m_{est,i})^T}{I}
\end{equation}
\normalsize

where $W_a$, $\bar{r}^m_{est} \in \mathbb{R}^{N \times 6}$, $\bar{r}_{mea} \in \mathbb{R}^{N \times 6}$, and $I$ are the selected model, $m^{th}$ model's estimation, payload residual measurements, and the number of data frames of the calibration trajectory. $W_a$ will remain activated until the next calibration.

\vspace{-4pt}
\subsection{Payload-adaptive Pre-trained Model (PaPM)}

\begin{figure}[t]
    \centering
    \includegraphics[width=0.47\textwidth]{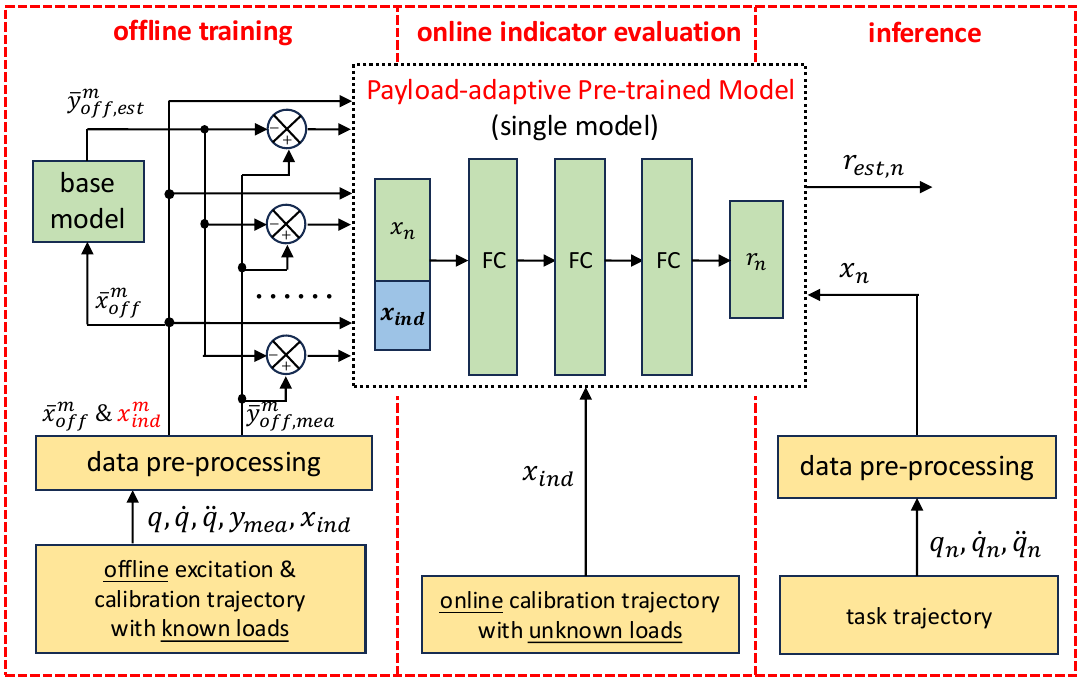}
    \vspace{-2pt}
    \caption{The block diagram and data flow for PaPM. The implication of the variables can be found in Table \ref{Table:Var_Imp}.}
    \label{Fig:IndLD}
    \vspace{-10pt}
\end{figure}

Although PsPM can provide highly accurate payload compensation across the entire joint space, loading all models into the system could significantly increase memory usage. Furthermore, with fine-grained discretization of the payload, resulting in numerous models being tested online, the computation time required for both inference and model selection becomes substantial. For example, the computation time for equations (5) and (6) across 69 models on 400 data frames (4 seconds) exceeds 2 seconds, despite parallel computation. In response to these challenges, we propose the PaPM in this section. The data flow and block diagram of this approach are depicted in Fig. \ref{Fig:IndLD}. PaPM is trained with all the data collected for PsPM but incorporates a unique element, the Payload Indicator (PI, $x_{ind}$), obtained from a carefully designed calibration trajectory, to distinguish between PsPM datasets.

The format and length of the PI vector $x_{ind}$ can be flexible, as long as it effectively reflects the unique identity of the payload. For example, it can include the magnitude of the mismatch between loaded measurements and load-free estimations. In practice, the PI vector should be designed in conjunction with the calibration trajectory, as its format and length may vary significantly with different trajectories. Theoretically, in a non-singular configuration, a single current measurement uniquely correspond to an end-effector payload and is sufficient for evaluating the PI vector. However, disturbances during low-speed operations - unnecessarily white noise \cite{zhang2021modeling} - may lead to inconsistent measurements across multiple calibration trials of the same payload. In contrast, evaluating the PI vector from a trajectory helps reduce the uncertainty of single measurements and ensures the acquisition of the same values in every trial. Section IV.B will detail our implementation for the PI vector. 

For capturing payload-specific information, the same calibration trajectory is executed alongside the standard excitation trajectories during PsPM data collection. It is consistent for different payloads and is executed for each new payload dataset collected. In light of the objective for fast calibration, two key guidelines must be adhered to in the design of the calibration trajectory: (i) The trajectory should have an execution time of less than 5 seconds. (ii) It should effectively convey the payload information at the joint level. Specifically, it should avoid singularities and positions near singularities, where payload variations result in no or negligible changes to the current measurements of specific joints. Section IV.B will detail our implementation for the calibration trajectory. 

Similar to PsPM, PaPM employs a 3-layer fully-connected MLP architecture, trained to predict payload residuals. During the training phase, each input data frame $x^m_n$ is concatenated with the PI vector $x^m_{ind}$ to form the complete PaPM input. Notably, the PI vector is disturbed with white noise to mimic noisy measurements and prevent the model from making out-of-distribution estimations when encountering unseen PI vectors during inference. The variance of the white noise is set at 0.25, taking into account the discretization resolution.  For online calibration, the designated calibration trajectory is executed to evaluate the PI vector for the unknown payload. This PI vector is then incorporated into the input for applications and remains unchanged until the next calibration.

\begin{table}[t]
    \begin{tabularx}{0.48\textwidth}{|>{\centering\arraybackslash}m{.12\linewidth}|X|}
        \hline
        Symbol & \multicolumn{1}{c|}{Implications}  \\ \hline 
        \multirow{2}{*}{$x_n$} & Real-time input vector of the $n^{th}$ frame,
        comprising the joint states and Motion Discriminators (MD) in \cite{shan2024sensorless}. \\ \hline 
        \multirow{2}{*}{$y_{est,n}$} & Real-time current estimation of the
        $n^{th}$ frame using the base or fine-tuned base model \\ \hline 
        $\bar{x}$ & Stacked input vectors  \\ \hline
        $\bar{y}_{mea}$ & Stacked current measurement vectors \\ \hline
        $q, \dot{q}, \ddot{q}$ & Joint states including position, velocity, and acceleration \\
        \hline
        $q_n, \dot{q}_n, \ddot{q}_n$ & Real-time joint states of the $n^{th}$ frame \\
        \hline
        \multirow{2}{*}{$\bar{x}^m_{off}$} & Stacked input vectors for offline
        training collected using the $m^{th}$ payload \\ \hline
        \multirow{2}{*}{$\bar{y}^m_{off,...}$} & Stacked current measurements or
        base model estimations involved in offline training collected with the
        $m^{th}$ payload \\ \hline
        $\bar{r}^m_{est}$ & Stacked payload residual estimation of the $m^{th}$
        model \\ \hline
        $\bar{r}_{mea}$ & Stacked payload residual measurements \\ \hline
        $r_{est,n}$ & Real-time payload residual estimation of the $n^{th}$
        frame \\
        \hline
        $x_{ind}$ & Payload Indicator (PI) vector \\ \hline
    \end{tabularx}
    \caption{Implications of variables presented in Fig. \ref{Fig:OnLD}, \ref{Fig:OffLD}, \ref{Fig:IndLD}}
    \label{Table:Var_Imp}
    \vspace{-6pt}
\end{table}

\section{Training Data Collection}

This section provides implementation details of the data collection procedure, excitation trajectories, and calibration trajectory. The robot manipulator used is the Denso-VS060, a position-controlled robot without force/torque sensing capabilities. The robot's hardware interface, motion planning algorithms, and data acquisition algorithms were all developed within the Robot Operating System (ROS) framework.

\subsection{Data Collection for Offline Learning}

\subsubsection{Payload Discretization}

Through careful design of the end-effector, it is possible to adjust payload variables such as mass, Center of Mass (CoM), and inertia. Given that inertia plays a minor role in the low-speed tasks of interest and is challenging to adjust systematically, we focus exclusively on mass and CoM as the key payload variables.

\begin{figure}[t]
    \centering
    \includegraphics[width=0.33\textwidth]{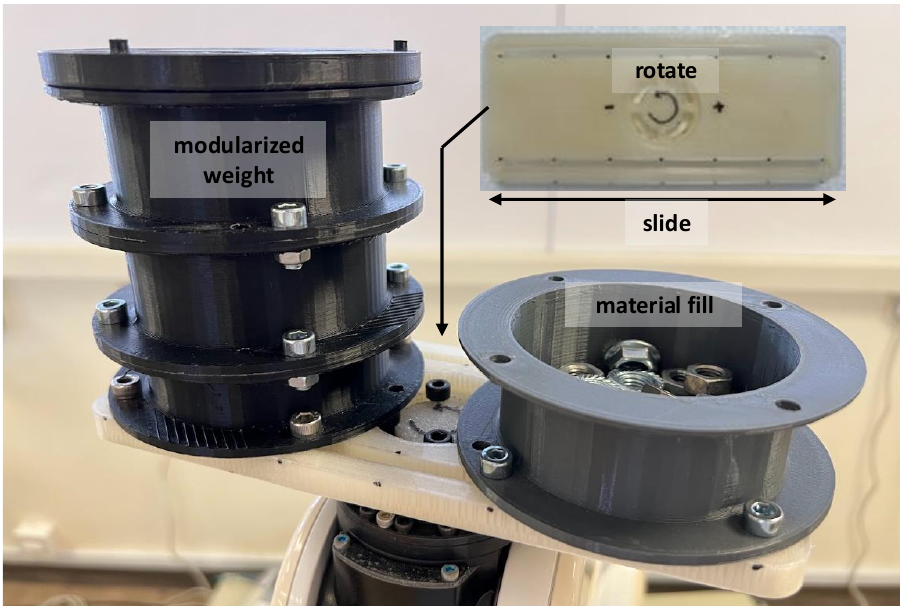}
    \caption{The end-effector design for payload adjustment.}
    \label{Fig:EE_design}
    \vspace{-12pt}
\end{figure}

The end-effector design is illustrated in Fig. \ref{Fig:EE_design}. This design facilitates the adjustment of mass and CoM by filling the containers with various materials, enabling payload discretization. Most end-effector and gripper designs are geometrically symmetrical around the Z-axis, thus assuming a centric CoM ($c_Lx=0$, $c_Ly=0$) with fine-grained discretization of $m$ and $c_{Lz}$ could be beneficial for efficient training data collection, meeting the requirements for most applications. Nevertheless, to demonstrate the robustness and capability of the proposed method in learning from diverse payloads, we collected off-centric training data with rough discretization. Table \ref{Table:payload} presents 21 sets of centric variables and 48 sets of off-centric variables, totaling 69 data sets, for training PsPM and PaPM. 

\begin{table}[t]
    \renewcommand{\arraystretch}{1.1}
    \centering  
    \setlength{\tabcolsep}{5.3pt}
    \begin{tabular}{|c|c c c c c c c|}
        \hline
        \multicolumn{8}{|c|}{Centric Payload Sets ($c_{Lx}=0$, $c_{Ly}=0$)} \\ \hline
    
        mass (kg) & 0.33  & 0.33  & 0.39  & 0.39  & 0.51  & 0.56  & 0.56  \\
        $c_{Lz}$ (m) & 0.035 & 0.060 & 0.082 & 0.096 & 0.060 & 0.085 & 0.103 \\ \hline

        mass (kg) & 0.73  & 0.88  & 0.91  & 0.91  & 1.05  & 1.15  & 1.15  \\ 
        $c_{Lz}$ (m) & 0.050 & 0.061 & 0.075 & 0.098 & 0.125 & 0.143 & 0.172 \\ \hline

        mass (kg) & 1.21  & 1.48  & 1.75  & 1.75  & 2.21  & 2.45  & 2.73  \\ 
        $c_{Lz}$ (m) & 0.091 & 0.129 & 0.079 & 0.070 & 0.115 & 0.124 & 0.129 \\ \hline
        \multicolumn{8}{c}{} \\ 
    \end{tabular}
    
    \vspace{-5pt}
    \begin{tabular}{|c|c c c c c c|}
        \hline
        \multicolumn{7}{|c|}{Off-centric Payload Sets for Each ($m,c_{Lz}$) in} \\ 
        \multicolumn{7}{|c|}{$\{(0.730,0.050),(1.083,0.087),(1.483,0.107)\}$} \\\hline
        $c_{Lx}$ (m) & -0.055 &  0.052 & -0.030 & -0.026 & -0.025 & -0.020 \\
        $c_{Ly}$ (m) &  0.000 & -0.030 &  0.052 & -0.015 &  0.000 &  0.035 \\ \hline
        $c_{Lx}$ (m) &  0.000 &  0.000 &  0.000 &  0.000 &  0.010 &  0.017 \\ 
        $c_{Ly}$ (m) & -0.055 & -0.030 &  0.020 &  0.050 & -0.019 &  0.010 \\ \hline
        $c_{Lx}$ (m) &  0.025 &  0.030 &  0.043 &  0.060 &        &        \\ 
        $c_{Ly}$ (m) & -0.048 &  0.000 &  0.025 &  0.000 &        &        \\ \hline
    \end{tabular}
    
    \caption{Discretized mass and CoM of centric and off-centric payloads used for data collection.} \label{Table:payload}
    \vspace{-12pt}
\end{table}

\subsubsection{Trajectory Excitation}

Sinusoidal exciting trajectories are commonly used for inverse dynamics identification. We refer to \cite{gaz2019dynamic} for the following trajectory expression:

\vspace{-6pt}
\small
\begin{equation}    
    q_j(t) = \sum_{l=1}^L \frac{a_{l,j}}{l\omega_f}sin(l\omega_ft) - \frac{b_{l,j}}{l\omega_f}cos(l\omega_ft) + q_{0,j}
    \vspace{-3pt}
\end{equation}
\normalsize
where $L=5$ represents the number of harmonics, $\omega_f = 0.15\pi$ in the original design, and the coefficients $a_{l,j}$, $b_{l,j}$, and $q_{0,j}$ are pre-determined constants, carefully chosen to cover the entire joint space. However, two factors make this specific trajectory less suitable for our learning-based method. First, Neural Network training typically requires a substantial amount of data. The 21-second trajectory in the original study, despite having manually selected optimal coefficients, does not provide sufficient dynamic information for NN-based methods. Second, the described trajectory is continuous. Previous research \cite{liu2021sensorless2} has shown that hysteresis becomes a significant issue with stationary joints or those in the pre-sliding zone. This concern is overlooked by continuous trajectories.

To this end, we introduce interruption-rich exciting trajectories for long-duration data collection. Given the sinusoidal nature of the trajectory, merely extending the execution time would result in repetitive robot motion. To ensure at least 30 minutes of varied trajectories for each payload, we generated multiple trajectories by following the steps below:
\begin{enumerate}[\hspace{0pt}(1)]
    \item For each trajectory, randomly sampled the coefficients $a_{l,j}$, $b_{l,j}$, and $q_{0,j}$ in equation (7) within the ranges of [-0.5, 0.5], [-0.5, 0.5], and the joint space, respectively. 
    \item Modified $\omega_f$ to scale down the velocity to better align with the low-speed requirements of tasks.
    \item Introduced interruptions into the continuous trajectories. Specifically, a pause of $t_p$ seconds is introduced after every $t_e$ seconds of execution, with $t_e$ and $t_p$ randomly sampled from the ranges of [1.0 s, 3.0 s] and [7.0 s, 9.0 s], respectively. Notably, such an interruption was applied to different joints separately. 
\end{enumerate}

Given that the Denso VS060 is a position-controlled robot, it provides only joint positions as the state measurements. Consequently, velocities and accelerations were obtained from the first and second-order derivatives of the position data. A Butterworth Filter: 
\vspace{-8pt} 
\begin{equation}
    \begin{split}
    v_n &=  \sum_{t=0}^{3}b_tu_{n-t} - \sum_{t=1}^{3}a_tv_{n-t} \\
    [a_1, a_2, a_3] &= [-2.592, 2.264 ,-0.664] \\
    [b_0, b_1, b_2, b_3] &= [0.0009, 0.0026, 0.0026 ,0.0009]
    \end{split}
\end{equation}
where $u$ and $v$ are the filter input and output, was applied to reduce signal noise, featuring a cutoff frequency of approximately 7 Hz under a 200 Hz sampling rate.  Joint positions, velocities, accelerations, and currents were recorded at a system frequency of 100 Hz.

\subsubsection{Data Size Investigation}

With the PsPM training scheme, we conducted preliminary experiments to explore the relationship between estimation errors ($e$ = $y_{mea} - y_{est} - r_{est}$) and the volume of training data. We collected the sample dataset over two hours of excitation trajectories executed with a consistent payload ($m$ = $0.91$ kg, $c_{Lz}$ = $0.075$ m). Models were then trained using different portions of this dataset. Details on the model size and training parameters are provided in Section V.

\begin{figure}[t]
    \centering
        \minipage{0.28\textwidth}
            \includegraphics[width=\linewidth]{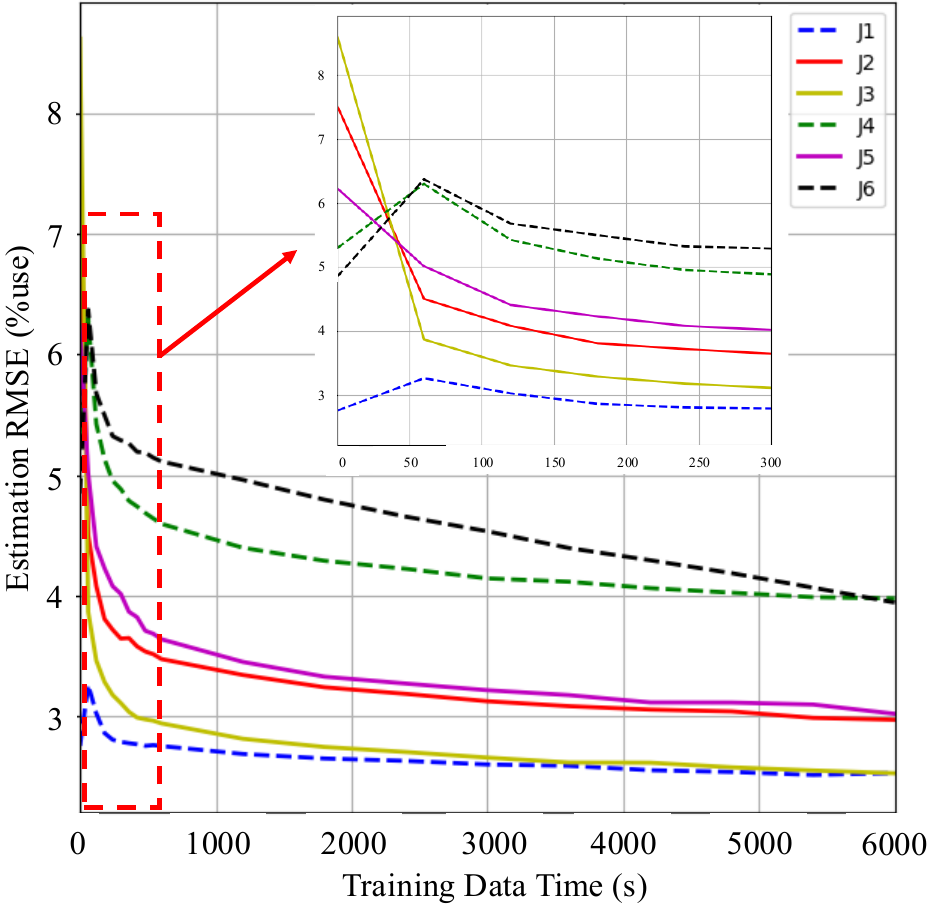}
        \endminipage\hfill
        \minipage{0.18\textwidth}
            \centering
            \includegraphics[width=\linewidth]{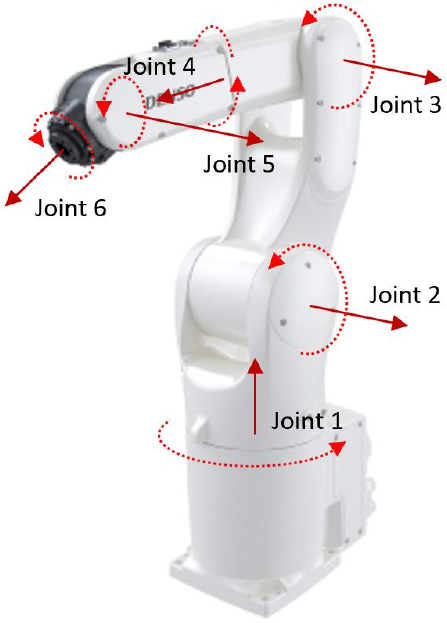}
        \endminipage\hfill
        \caption{The estimation error for each joint plotted against the training data length, and Denso VS060 labeled with axises.}
        \label{Fig:DataSizeInv}
        \vspace{-10pt}
\end{figure}

Fig. \ref{Fig:DataSizeInv} displays the estimation errors of all models, tested on a different dataset that covers a wide joint space, and plotted in relation to the length of the training data. Clearly, a larger training set size leads to reduced estimation errors, although the benefits diminish for training durations exceeding 30 minutes. In light of this diminishing return, we collect a 30-minute dataset for each discretized payload. Longer trajectories could be considered if time constraints for data collection are relaxed, or if higher accuracy is required. Our design resulted in 180,000 (30 mins) data frames for each PsPM dataset, cumulatively amounting to 12,420,000 (34.5 hours) data frames for PaPM training.

Interestingly, within the segment of Fig. \ref{Fig:DataSizeInv} that spans from 0 to 300 seconds,  we observe two distinct patterns of error variation. The solid lines represent joints 2, 3, and 5, which are significantly influenced by the end-effector payload due to the robot's geometry. Training the calibration model with merely 60 seconds of data proves to effectively reduce errors for these joints. In contrast, the dashed lines represent joints 1, 4, and 6, which are less affected by the payload. For these joints, the estimation error initially increases with smaller datasets and then decreases as the dataset size expands. This pattern indicates that when the calibration data is insufficient, the NN model tends to overfit the robot dynamics in a local region, resulting in decreased accuracy across the entire joint space. This observation also elucidates why the model-based method and OLM exhibit lower accuracy, which will be discussed in Section V.
\vspace{-10pt}

\subsection{Implementation of PI and Calibration Trajectory}

\begin{figure}[t]
    \centering
    \includegraphics[width=0.45\textwidth]{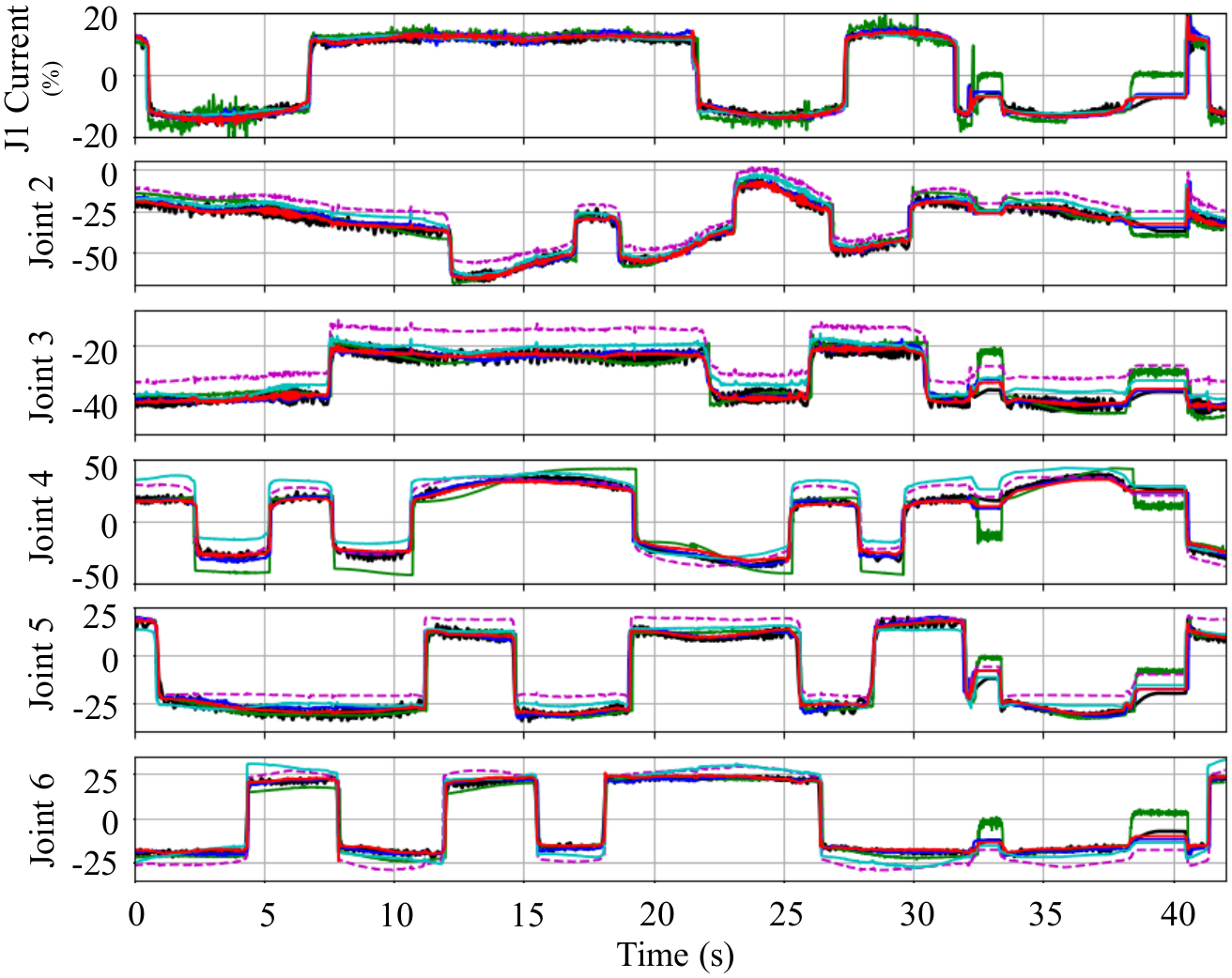}
    \vspace{-4pt}
    \caption{Comparison of the proposed methods with the baseline. The black, dashed magenta, green, cyan, blue, and red lines indicate respectively the measurements, uncalibrated base model, model-based method, OLM, PsPM+base, and PaPM+base. }
    \label{Fig:MethodComp}
    \vspace{-12pt}
\end{figure}

In accordance with the concept discussed in Section III.D, our implementation of the calibration trajectory comprises two steps: (1) selecting a static position distal from singularities to maximally distinguish payloads, and (2) commanding all joints to move locally around the selected position to acquire reliable PI values while avoiding hysteresis errors. Specifically, we designed a 4-second calibration trajectory for online model selection (PsPM) and PI evaluation (PaPM). The robot's initial position was set at [0$^\circ$,40$^\circ$,50$^\circ$,45$^\circ$,45$^\circ$,0$^\circ$]. For the first 2 seconds, all joints maintain constant speeds of [3$^\circ$/s,3$^\circ$/s,3$^\circ$/s,-3$^\circ$/s,-3$^\circ$/s,-3$^\circ$/s], with these velocities reversing in the final 2 seconds. The command velocity was doubled for the model-based method and OLM to span larger joint space. The calibration algorithms for OLM and PsPM are discussed sufficiently in Section III.A and III.B. In the following, we will discuss the evaluation details of the PI vector in our setting.

The PI vector $x_{ind} \in \mathbb{R}^{12 \times 1}$ was obtained by concatenating two vectors $x^1_{ind} \in \mathbb{R}^{6 \times 1}$ and $x^2_{ind} \in \mathbb{R}^{6 \times 1}$, which were the average payload residuals measured in the first and second half of the calibration trajectory, respectively. Both $x^1_{ind}$ and $x^2_{ind}$ were evaluated using the following equation:
\vspace{-3pt}
\begin{equation}
    x^k_{ind} = \frac{\sum_{n=1}^N y_{mea,n} - y_{est,n}}{N}, \qquad k=1,2
\end{equation}
where $y_{mea} \in \mathbb{R}^{6 \times 1}$ is the measured currents, $y_{est} \in \mathbb{R}^{6 \times 1}$ is the base-model-estimated current, and $N$ is the number of frames recorded for each 2-second trajectory. 

\section{Model Training and Comparisons}

The fully-connected hidden layers in all models described in Sections III and IV, including OLM, PsPM, PaPM, and the preliminary test model, consist of 512 neurons and utilize the ReLU activation function. For PsPM and PaPM, the PyTorch framework was employed for model training during the offline learning phases, and all models were implemented using the Python-enabled NumPy library for online inference. During training, $MSEloss$ and $Adam$ regularization were applied, along with a set of hyperparameters: a 0.5 random dropout rate, a training batch size of 4096, and a learning rate of 0.001. For OLM, fine-tuning the base model adheres to the same training procedures. However, the 4-second calibration trajectory produces only 400 data frames, allowing for the model's fine-tuning to be completed within 3 seconds.

For the qualitative evaluation of calibration outcomes, a test trajectory was executed both before and after the calibration process. Since PsPM and PaPM predict only payload residuals, their estimations offset from the base model's estimations. Fig. \ref{Fig:MethodComp} shows the estimation results for the test trajectory, comparing the performance of the model-based method, OLM, PsPM, and PaPM. These methods significantly reduce estimation errors in joints 2, 3, and 5, which were notably affected by the payload ($m$ = $0.83$ kg, $c_L$ = [$0.000$ m, $0.030$ m, $0.087$ m]). However, for the model-based method and OLM, although the joint space explored was larger given the doubled velocity, the data collected online was still relatively local, leading to inaccurate estimations when the Inverse Kinematics Class was distal from that of the calibration trajectory. The Root Mean Square Errors (RMSE) for 90,000 test data frames are detailed in Table \ref{Table:RMSE}. These results suggest that PsPM and PaPM both deliver high accuracy across various configurations. Given the advantages of PaPM outlined in Section III, PaPM was preferred over PsPM for the applications in Section VI.

To further validate the robustness of PaPM, we conducted additional tests using five different PaPMs, each corresponding to a unique calibration trajectory across the joint space, as well as four different payloads, all on the same test trajectory. Table \ref{Table:RMSE} presents the average results of these tests.

\vspace{-3pt}

\begin{table}[t]
    \centering  
    \begin{tabular}{|c|c c c c c c|}
        \hline
                 & \multicolumn{6}{c|}{RMSE on Test Trajecotry (\%use)} \\ \hline
                 & J1    & J2    & J3     & J4    & J5     & J6        \\ \hline
        Base     & 1.52  & 8.36  & 10.05  & 7.38  & 8.93  & 4.97      \\
        Model-based \cite{gaz2017payload}  &  2.46  &  6.25  &  4.75  &  8.56  &  5.18  &  6.17 \\
        OLM      & 1.56  & 4.15  & 4.05   & 8.80  & 4.91  & 3.88      \\
        PsPM + Base    & 1.62  & 2.34   & 2.26   & 5.82  & 2.75  & \textbf{2.32} \\ 
        PaPM + Base    & \textbf{1.32}  & \textbf{2.09}   & \textbf{1.97}  & \textbf{4.03}  & \textbf{2.16}  & 2.87 \\ \hline
        & \multicolumn{6}{c|}{PaPM + Base RMSE on Robustness Tests} \\ \hline
        
        5 trajectory average & 1.53 & 2.20 & 1.97 & 4.25 & 2.34 & 2.70\\ 
        4 payloads average & 1.36 & 2.19 & 2.00 & 4.21 & 2.19 & 3.17\\ \hline
    \end{tabular}
    
    \vspace{5pt}
    \begin{tabular}{m{2.5cm} m{2.5cm} m{2.5cm}}
        \hline
        \multicolumn{3}{|c|}{Initial Positions of Tested Calibration Trajectories ($^\circ$)} \\ \hline 
        \multicolumn{3}{|c|}{\hspace{-3.5pt}[-40, 30, 30, 45, 20, -40]; [-20, 45, 0, -45, 45, 20];} \\ 
        \multicolumn{3}{|c|}{[0, 10, 55, 30, 60, 0]; [20, 20, 20, -45, 30, -20]; [40, 0, 40, 45, 50, -40]} \\ \hline 
    \end{tabular}

    \vspace{5pt}
    \begin{tabular}{m{2.5cm} m{2.5cm} m{2.5cm}}
        \hline
        \multicolumn{3}{|c|}{Parameters of Tested Payloads [$m$, $c_{Lx}$ $c_{Ly}$, $c_{Lz}$] (kg, m, m, m)} \\ \hline
        \multicolumn{3}{|c|}{[0.905, 0.000, 0.000, 0.075]; \hspace{1pt} [1.211, 0.019, 0.012, 0.079];} \\
        \multicolumn{3}{|c|}{[0.850, 0.000, -0.040, 0.108]; \hspace{1pt} [0.675, -0.025, 0.045, 0.045]} \\ \hline
    \end{tabular}

    \caption{The bold values indicate the smallest error. 'Base' indicates the base model without calibration.}
    \label{Table:RMSE}
    \vspace{-15pt}
\end{table}

\section{Application}

\subsection{Sensorless Joint Space Compliance}

In the joint compliance task, robots are expected to respond compliantly to external contacts with the human operator. This objective is often straightforwardly achieved through gravity compensation when users can directly command joint torques. However, the Denso-VS060 is a position-controlled robot, which prohibits direct torque or current commands. Consequently, an alternative method is necessary to enable sensorless joint compliance. Previous research \cite{de2006collision} has provided a comprehensive analysis of this challenge. An admittance controller:
\vspace{-6pt}
\begin{equation}
    \dot{q}_n = K_py_{ext,n} = K_p(y_{mea,n}-y_{est,n}-r_{est,n}) \\
    \vspace{-3pt}
\end{equation}
where $K_p$ and $y_{ext,n}$ are the constant gain and external residual estimation, respectively, is utilized to achieve compliant motions without the need for direct current commands. To prevent unexpected movements in the absence of contact, a deadzone was imposed on $y_{ext,n}$, ensuring a joint activates only in response to sufficiently large external forces.

To begin with, we compared the estimation error in two scenarios: base model estimation without a payload and PaPM-compensated estimation with a payload. Intuitively, the deadzone selection should be identical since PaPM only calibrates the payload. However, experimental results show that the calibration model also behaves as an additional model hierarchy, as presented in \cite{shan2024sensorless}. As a result, in addition to compensating for the payload residuals, PaPM further reduces the estimation error left by the base model. This enhancement has enabled the selection of smaller deadzone boundaries, making the robot feel more responsive and 'lighter'.  Details on the deadzone values are provided in Table \ref{Table:DDZ_bound}. Notably, expressing the current's unit in percentage usage (\%use) is less intuitive for force analysis. Therefore, equivalent torque values were approximated using measurements from the F/T sensor, taken at specific static postures and considering the robot's geometry.

\begin{table}[t]
    \centering 
    
    \begin{tabular}{|c|c c c c c c|}
        \hline
                            & \multicolumn{6}{c|}{Selection of Deadzone Boundary}      \\ \hline
                            & J1    & J2     & J3     & J4     & J5     & J6     \\ \hline
        Base \cite{shan2024sensorless} (\%use) & 6.0   & 6.0    & 6.0    & 9.0    & 11.0   & 11.0   \\
        Base+PaPM (\%use)    & 6.0   & 6.0    & 4.0    & 6.5    & 7.0    & 7.0    \\ \hline
        Motor const. est    & 0.71  & 0.81   & 0.40   & 0.11   & 0.14   & 0.08   \\ \hline
        Base \cite{shan2024sensorless} (Nm)    & 4.26  & 4.86   & 1.61   & 0.72   & 1.00   & 0.53   \\
        Base+PaPM (Nm)       & 4.26  & 4.86   & 1.07   & 0.52   & 0.64   & 0.34   \\ \hline
    \end{tabular}
    \caption{Comparisons of the the deadzone boundary values for the base model without payload, and base model+PaPM with payload. The estimated motor constants are also provided.}
    \vspace{-8pt}
    \label{Table:DDZ_bound}
\end{table}

\begin{figure}[t]
    \centering
    \begin{subfigure}{0.46\textwidth}
        \includegraphics[width=\linewidth]{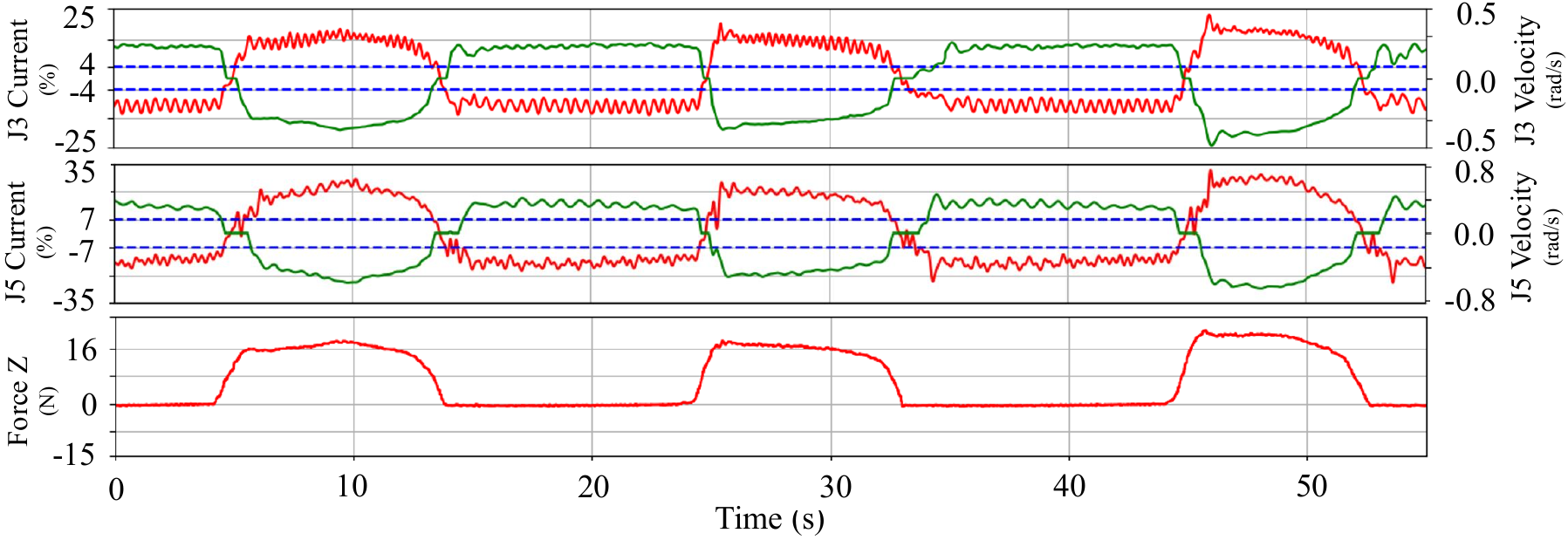}
        \vspace{-17pt}
        \caption{}
        \label{Fig:JC_no_cali}
        \vspace{-1pt}
    \end{subfigure}

    \begin{subfigure}{0.46\textwidth}
        \includegraphics[width=\linewidth]{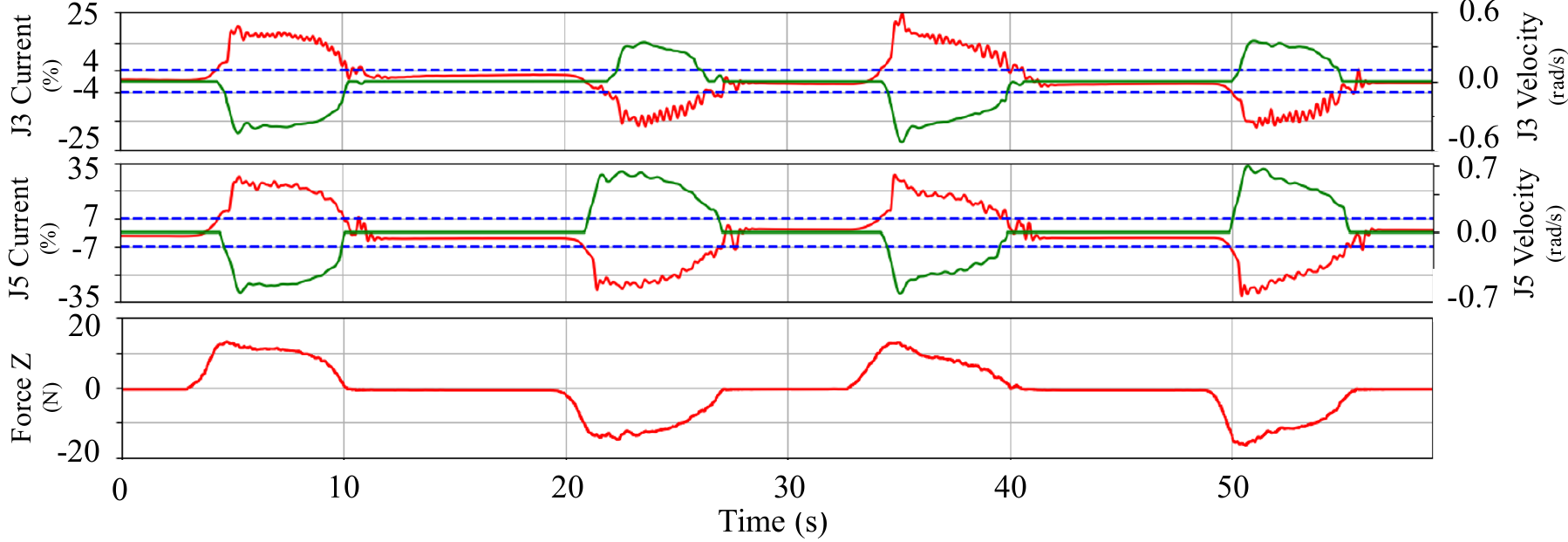}
        \vspace{-18pt}
        \caption{}
        \label{Fig:JC_with_cali}
    \end{subfigure}
    \vspace{-6pt}
    \caption{The joint compliance experiment (a) before (b) after calibration. In the top two plots, the red, green, and dashed magenta lines indicate the estimated residuals, command velocity, and deadzone boundaries. In the third plot, the red line indicates measured force in the Cartesian space.}
    \label{Fig:JC}
    \vspace{-10pt}
\end{figure}

In this experiment, the end-effector utilized has a mass and CoM of ($m$ = $0.83$ kg, $c_L$ = [$0.000$ m, $0.030$ m, $0.087$ m]).  The joint compliance mode enables whole-body compliance, as demonstrated in the supplementary video. In the subsequent experiment, the operator interacted only with the end-effector to record external contacts via the F/T sensor. Initially, the system was activated using the base model, without any calibration. Once joint compliance was enabled, the end-effector began to drop due to the gravity of the uncalibrated payload. This behavior is clearly illustrated in Fig. \ref{Fig:JC_no_cali}, where non-zero joint current residuals were observed, even though no force was detected by the sensor. In contrast, after calibration, Fig. \ref{Fig:JC_with_cali} displays a clear correlation between the measured force and the estimated residual, indicating that the payload was accurately calibrated. This allowed the robot to cease movements immediately upon the removal of external forces.

\vspace{-7pt}
\subsection{Wrench Estimation and Task Compliance}

External Wrench Estimation (WE) is another essential application of dynamics identification. It facilitates the detection and control of contact forces exclusively through the use of current and joint state measurements \cite{wahrburg2017motor}, enabling complex industrial tasks such as high-precision assembly, without the necessity for external sensing devices.

\begin{figure}[t]
    \centering
    \includegraphics[width=0.43\textwidth]{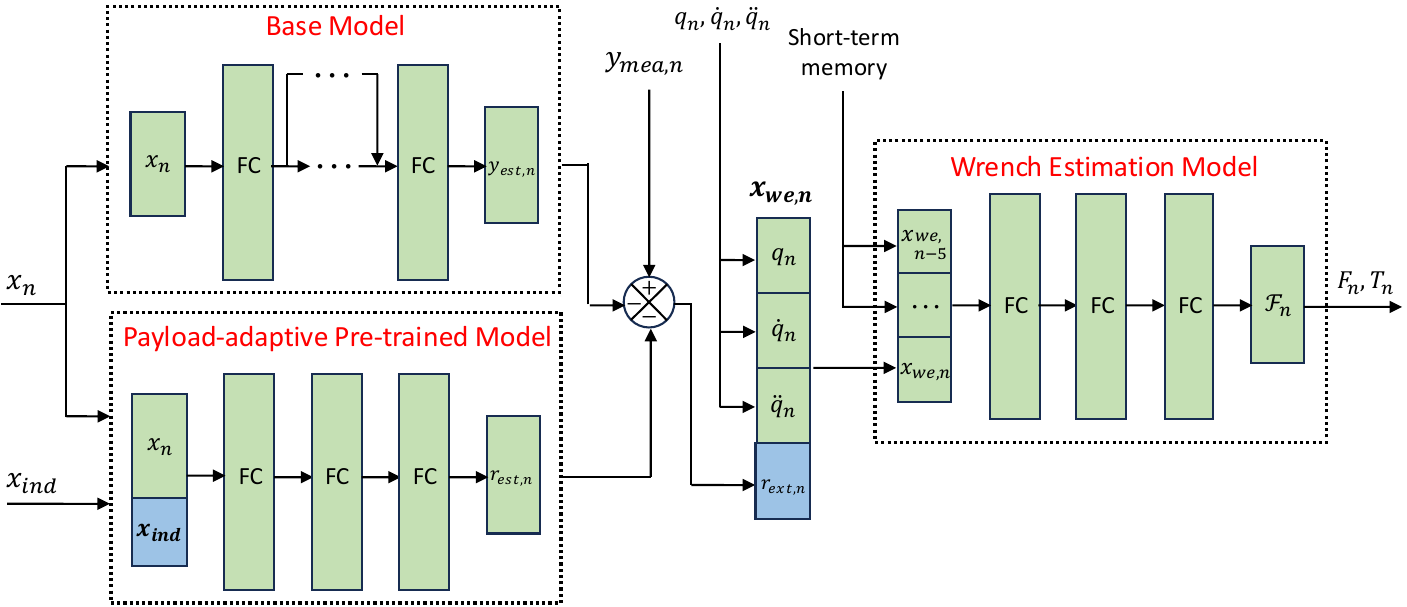}
    \caption{The block diagram of the wrench estimation scheme. $x_{we,n}$ indicates the input vector unit including the joint states and external force residuals of the $n^{th}$ frame.}
    \label{Fig:WED}
    \vspace{-15pt}
\end{figure}

Due to the absence of exact motor constants, directly determining the end-effector wrench using the estimated current residual and the robot's Jacobian Matrix is not feasible. To address this challenge, we draw upon methodologies from previous research \cite{10354486}, which introduces a systematic data collection procedure for wrench estimation, and \cite{shan2024sensorless}, which implemented compound NN models. The conceptual framework of this approach is depicted in Fig. \ref{Fig:WED}. The WE model employs a fully-connected MLP architecture. It inputs the instantaneous joint state vector $(q_n, \dot{q}_n, \ddot{q}_n)$ along with an estimation of the payload residuals $r_{est,n}$. Additionally, the model incorporates a short-term memory mechanism, utilizing the vector ($x_{we,n-1},...,x_{we,n-5}$) from the preceding five frames to enhance accuracy and minimize noise.

After training the WE model, it was integrated into the ROS framework to facilitate real-time wrench estimation. For task space motion control, an admittance controller was employed, leveraging the wrench estimates instead of direct measurements. Subsequent joint space motion was determined using Inverse Kinematics. Additionally, a deadzone was carefully selected to set thresholds for initiating motion.

\begin{figure}[t]
    \centering
    \includegraphics[width=0.46\textwidth]{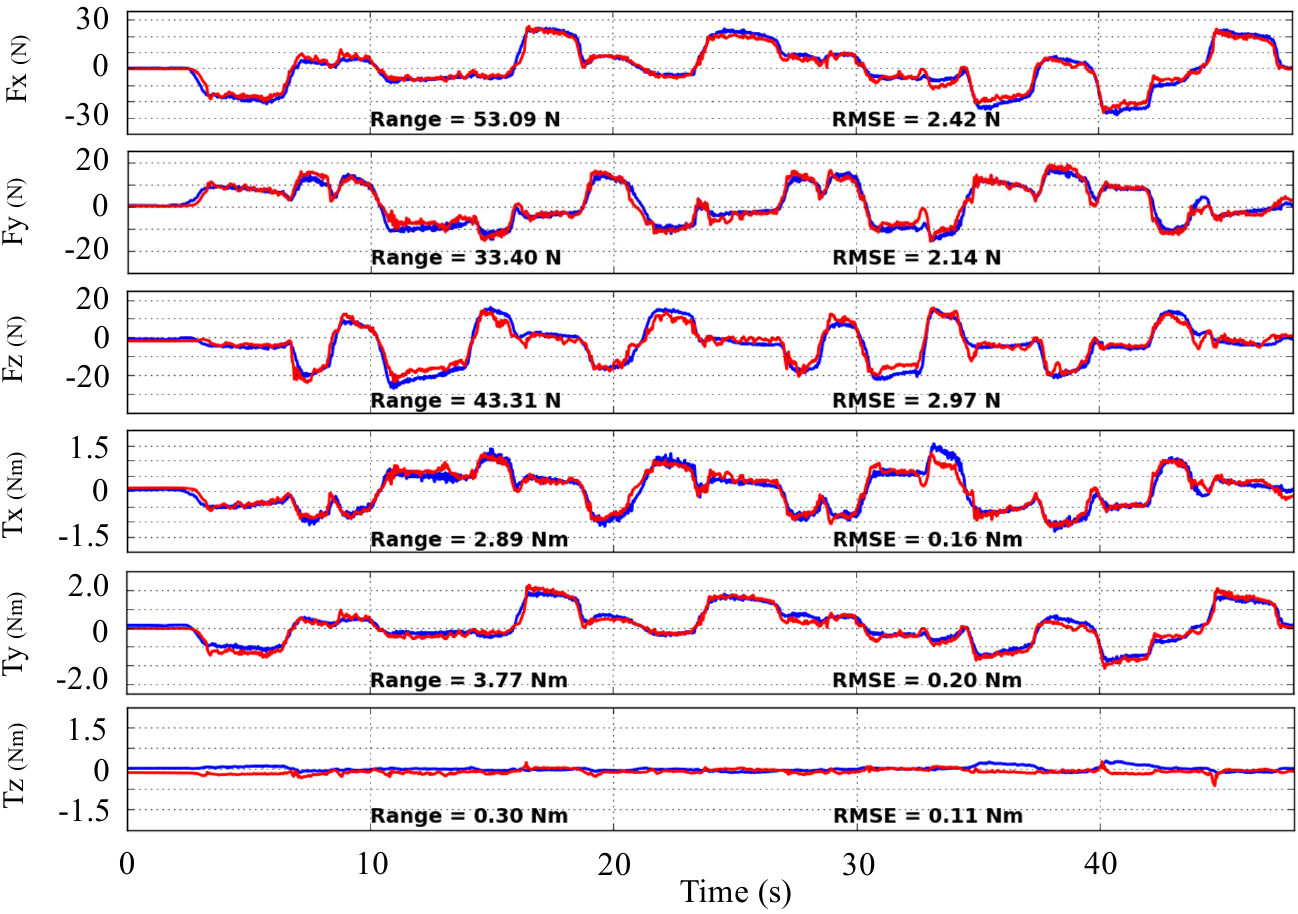}
    \vspace{-5pt}
    \caption{Comparisons of the wrench estimations (red) and measurements (blue) in task compliance after online calibration using PaPM. The weight of payload is compensated for wrench measurements.}
    \label{Fig:TC}
    \vspace{-7pt}
\end{figure}

In this experiment, after pre-calibrating the weight of the end-effector, which included a coupled F/T sensor and gripper, the robot's task compliance mode was activated. The robot then proceeded to grasp an aluminum block (0.62 kg), which was treated as an unknown payload for the purpose of calibration. Upon grasping the aluminum block, a constant offset in the estimated wrench was observed, leading to a gradual descent of the end-effector until it reached the boundary. This behavior is displayed more clearly in the supplementary video. Subsequently, the same 4-second calibration was conducted. The results of this calibration, both quantitative and qualitative, are presented in Fig. \ref{Fig:TC}. Overall, the calibration's accuracy is evidenced by nearly overlapped curves between the estimations and measurements, along with minimal RMSE.

\section{Conclusion}

In this study, we presented a fast end-effector payload calibration technique for dynamics identification and its application in joint space compliance and wrench estimation. This approach employs Neural Networks to model robot dynamics accurately across various payload scenarios, covering the entire joint space. Leveraging an extensive offline learning process, this method requires only minimal online data, enabling the use of short calibration trajectories. This outcome is particularly valuable in industrial settings where manipulators grasp a wide range of objects.

While the proposed calibration scheme achieves high levels of accuracy, learning-based methods generally rely on substantial volumes of training data, posing a challenge for easy replication on other robots. Additionally, the current payload discretization approach and end-effector design allow only adjustments of mass and CoM. Consequently, our implementation is limited to low-speed operation and collaboration, where the effect on payload inertia is less obvious. For replication on different robot models, it would be beneficial to develop a better payload discretization approach, such as an end-effector design that allows systematic adjustment of inertia. This would expand training data diversity while reducing the data size requirement.

Another promising avenue for applying the proposed method is with torque-controllable robots. For position-controlled robots, joint compliance should be implemented with careful selection of deadzones. Moreover, given the admittance control scheme, robot motion may be jerky if the current measurement is noisy. However, with torque/current-controllable robots, the estimated payload residuals can be integrated directly into gravity compensation. The quality of robot behaviors would no longer depend on deadzone selection and measurement noise, resulting in lighter, more responsive robot behaviors with smoother trajectories during interaction. 

\bibliographystyle{ieeetr}
\bibliography{reference} 

\end{document}